\documentclass[11pt,a4paper]{article}
\usepackage[hyperref]{eacl2021}
\usepackage{times}
\usepackage{latexsym}

\usepackage{natbib}
\usepackage{amsmath} 
\usepackage{caption} 
\usepackage{hyperref}       
\usepackage{url}            
\usepackage{xcolor}
\usepackage{booktabs}       
\usepackage{amsfonts}       
\usepackage{nicefrac}       
\usepackage{microtype}      
\usepackage{algpseudocode}
\usepackage{float}
\usepackage{caption}
\usepackage{subcaption}
\usepackage{graphics}
\usepackage{amssymb,graphicx}
\usepackage{times}  
\usepackage{helvet} 
\usepackage{courier}  
\usepackage{graphicx} 

\usepackage{booktabs}
\usepackage{makecell} 
\usepackage{multirow}
\usepackage{multicol}
\usepackage{algorithm}
\usepackage{adjustbox}
\usepackage{float}
\usepackage{caption}
\usepackage{subcaption}
\usepackage{graphics}
\newsavebox{\tempbox}
\usepackage{amsmath,amsthm,amssymb}
\usepackage{mathtools}
\usepackage{xspace}

\usepackage{microtype}

\aclfinalcopy 
\def\aclpaperid{***} 

\title{Annealing Knowledge Distillation}

\author{\textsuperscript{1,2}Aref Jafari, \textsuperscript{2}Mehdi Rezagholizadeh, \textsuperscript{1}Pranav Sharma, \textsuperscript{1,3}Ali Ghodsi\\ \\
  \textsuperscript{1} David R. Cheriton School of Computer Science, University of Waterloo\\
  \textsuperscript{2}Huawei Noah's Ark Lab\\
  \textsuperscript{3}Department of Statistics and Actuarial Science, University of Waterloo\\
  \texttt{\{aref.jafari, p68sharma, ali.ghodsi\}@uwaterloo.ca}\\
  \texttt{mehdi.rezagholizadeh@huawei.com}
  \\}
\date{}

\begin{document}
\maketitle
\begin{abstract}
Significant memory and computational requirements of large deep neural networks restrict their application on edge devices. Knowledge distillation (KD) is a prominent model compression technique for deep neural networks in which the knowledge of a trained large teacher model is transferred to a smaller student model.
The success of knowledge distillation is mainly attributed to its training objective function, which exploits the soft-target information (also known as ``dark knowledge") besides the given regular hard labels in a training set. 
However, it is shown in the literature that the larger the gap between the teacher and the student networks, the more difficult is their training using knowledge distillation. To address this shortcoming, we propose an improved knowledge distillation method (called Annealing-KD) by feeding the rich information provided by the teacher's soft-targets incrementally and more efficiently. 
Our Annealing-KD technique is based on a gradual transition over annealed soft-targets generated by the teacher at different temperatures in an iterative process, and therefore, the student is trained to follow the annealed teacher output in a step-by-step manner. 
This paper includes theoretical and empirical evidence as well as practical experiments to support the effectiveness of our Annealing-KD method.
We did a comprehensive set of experiments on different tasks such as image classification (CIFAR-10 and 100) and NLP language inference with BERT-based models on the GLUE benchmark and consistently got superior results.
\end{abstract}

\section{Introduction}
Despite the great success of deep neural networks in many challenging tasks such as natural language processing~\citep{vaswani2017attention, liu2019roberta}, computer vision~\citep{wong2019yolo,howard2017mobilenets}, and speech processing~\citep{chan2016listen,he2019streaming}, these state-of-the-art networks are usually heavy to be deployed on edge devices with limited computational power~\citep{bie2019fully, lioutas2019distilled}. A case in point is the BERT model~\citep{devlin2018bert} which can be comprised of more than a hundred million parameters. 

The problem of network over-parameterization and expensive computational complexity of deep networks can be addressed by neural model compression. 
There are abundant of neural model compression techniques in the literature~\citep{prato2019fully,tjandra2018tensor,jacob2018quantization}, among which knowledge distillation (KD) is one of the most prominent techniques~\citep{hinton2015distilling}. 
KD is tailored a lot to serve different applications and different network architectures~\citep{furlanello2018born, gou2020knowledge}. For instance, patient KD~\citep{sun2019patient}, TinyBERT~\citep{jiao2019tinybert}, and MobileBERT~\citep{sun2020mobilebert} are designed particularly for distilling the knowledge of BERT-based teachers to a smaller student.   

The success of KD is mainly attributed to its training objective function, which exploits the soft-target information (also known as ``dark knowledge") besides the given regular hard labels in the training set~\citep{hinton2012neural}. 
Previous studies in the literature \citep{lopez2015unifying, mirzadeh2019improved} show that when the gap between the student and teacher models increases, training models with KD becomes more difficult. We refer to this problem as KD's \textit{capacity gap problem} in this paper.   
For example, \citet{mirzadeh2019improved} show that if we gradually increase the capacity of the teacher, first the performance of student model improves for a while, but after a certain point, it starts to drop. 
Therefore, although increasing the capacity of a teacher network usually boosts its performance, it does not necessarily lead to a better teacher for the student network in KD. In other words, it would be more difficult for KD to transfer the knowledge of this enhanced teacher to the student. 
A similar scenario happens when originally the gap between the teacher and student network is large. 

\citet{mirzadeh2019improved} proposed their TAKD solution to this problem which makes the KD process more smooth by filling the gap between the teacher and student networks using an intermediate auxiliary network (referred to as ``teacher assistant"). The size of this TA network is between the size of the student and the teacher; and it is trained by the teacher first. Then, the student is trained using KD when the TA network is playing the role of its teacher. This way, the training gap (between the teacher and the student) would be less significant compared to the original KD. However, TAKD suffers from the high computational complexity demand since it requires training the TA network separately. Moreover, the training error of the TA network can be propagated to the student during the KD training process. 

In this paper, we want to solve the KD \textit{capacity gap problem} from a different perspective.
We propose our Annealing-KD technique to bridges the gap between the student and teacher models by introducing a new KD loss with a dynamic temperature term. This way, Annealing-KD is able to transfer the knowledge of the teacher smoothly to the student model via a gradual transition over soft-labels generated by the teacher at different temperatures.
We can summarize the contributions of this paper in the following: 
\begin{enumerate}
    \item We propose our novel Annealing-KD solution to the KD \textit{capacity gap problem} based on modifying the KD loss and also introducing a dynamic temperature function to make the student training gradual and smooth.
    \item We provide a theoretical and empirical justification for our Annealing-KD approach. 
    \item We apply our technique to ResNET8 and plain CNN models on both CIFAR-10 and CIFAR-100 image classification tasks, and the natural language inference task on different BERT based models such as DistilRoBERTa, and BERT-Small on the GLUE benchmark and achieved the-state-of-the art results.  
    \item Our technique is simple, architecture agnostic, and can be applied on top of different variants of KD.
\end{enumerate}




\section{Related Work}

\subsection{Knowledge Distillation}
In the original Knowledge distillation method by~\citet{hinton2015distilling}, which is referred to as KD in this paper, the student network is trained based on two guiding signals: first, the training dataset or \textit{hard labels}, and second, the teacher network predictions, which is known as \textit{soft labels}. 
Therefore, KD is trained based on a linear combination of two loss functions: the regular cross entropy loss function between the student outputs and hard labels, and the KD loss function to minimize the distance between the output predictions of the teacher and student networks at a particular temperature, $\mathcal{T}$ , on training samples:  
\begin{equation}
\begin{split}
    & \mathcal{L} =(1-\lambda)\mathcal{L}_{CE} + \lambda \mathcal{L}_{KD} \\
    & \mathcal{L}_{CE} = H_{CE}\Big(y,(\sigma(z_s(x))\Big)  \\
    & \mathcal{L}_{KD} = \mathcal{T}^2 KL\Big(\sigma(\frac{z_t(x)}{\mathcal{T}}), \sigma(\frac{z_s(x)}{\mathcal{T}})\Big)
\end{split}
\label{eq:KD}
\end{equation}
where $H_{CE}(.)$ and $KL(.)$ are representing the cross entropy and KL divergence respectively,  $z_s(x)$ and $z_t(x)$ are the output logits from the student and teacher networks, $\mathcal{T}$ is the temperature parameter, $\sigma(.)$ is the softmax function and $\lambda$ is a coefficient between [0,1] to control the contribution of the two loss functions.
The above loss function minimizes the distance between the student model and both the underlying function and the teacher model assuming the teacher is a good approximation of the underlying function of the data. 

A particular problem with KD, that we would like to address in this paper, is that the larger the gap between the teacher and the student networks, the more difficult is their training using knowledge distillation \citep{lopez2015unifying,mirzadeh2019improved}.

\subsection{ Teacher Assistant Knowledge Distillation (TAKD)}
To address the capacity gap problem between the student and teacher networks in knowledge distillation, TAKD \citep{mirzadeh2019improved} proposes to train the student (of small capacity) with a pre-trained intermediate network (of moderate capacity) called teacher assistance. 
In this regard, we first train the TA with the {guidance} of the teacher network by using the KD method. Then, we can use the learned TA network to train the student network. Here, since the capacity of the TA network is between the capacity of the teacher and the student networks, therefore it can fill the gap between the teacher and student and enhance the complexity of the teacher and transfer its knowledge to the student network. 

As it is mentioned in \citep{mirzadeh2019improved}, a better idea could be using TAKD in a hierarchical way. So in this case, we can have several TAs with different levels of capacity from large capacities close to the teacher model to small capacities close to the student model. Then we could train these TAs consecutively from large capacities to small capacities in order to have a more smooth transfer of teacher's knowledge to the student model. But it will be difficult. Because, first, since we need to train a new model each time, it is computationally expensive. Second, in this way we will have additive error in each step. Each TA after training will have an approximation error and these errors will accumulate and transfer to the next TA. In the next section, we will propose a simple method to realize this idea and avoid the mentioned problems.

\subsection{Annealing in Knowledge Distillation}
{\citet{clark-etal-2019-bam} 
{proposed an annealing idea in their Born-Again Multi-task (BAM) paper}
, to train a multitask student network using distillation from some single-task teachers. 
They introduce a so-called teacher annealing scheme to distill from a dynamic weighted mixture of the teacher prediction and the ground-truth label. In this regard, the weight of teacher's prediction is gradually reduced compared to the weight of ground-truth labels during training.
Therefore, early in training, the student model mostly learns from the teacher and later on, it learns mostly from target labels.}
{However, our Annealing-KD is different from \citet{clark-etal-2019-bam} in different aspects. First, the introduced annealing term in BAM is conceptually different from our annealing. While in BAM, teacher annealing controls the contribution of the teacher dark knowledge compared to the ground-truth labels during training, our Annealing-KD is only applied to the teacher output in the KD loss to solve the capacity gap problem between the teacher and student networks. Second, the way we do annealing in our technique is through the temperature parameter and not {by controlling the contribution of the teacher and ground-truth labels}. Third, BAM falls into another category of knowledge distillation which focuses on improving the {performance} of the student model and not compressing it. Our method is described in the next section.  } 
\section{Method: Annealing Knowledge Distillation }
In this section, we describe our Annealing-KD technique and show the rationale behind it. 
First, we start by formulating the problem and visualizing our technique using an example for a better presentation. Then, we use VC-dimension theory to understand why our technique improves knowledge distillation. We wrap up this section by visualizing the loss landscape of Annealing KD for a ResNet network in order to investigate the impact of our method on the KD loss function. 

KD defines a two-objective loss function (i.e. the $\mathcal{L}_{KD}$ and $\mathcal{L}_{CE}$ terms in Equation~\ref{eq:KD}) to minimize the distance between student predictions and soft labels and hard labels simultaneously. {Without adding to the computational needs of the KD algorithm, our Annealing-KD model breaks the KD training into two stages}: Stage I, gradually training the student to mimic the teacher using our Annealing-KD loss $\mathcal{L}_{\text{KD}}^\text{Annealing}$; Stage II, fine-tuning the student with hard labels using $\mathcal{L}_{CE}$. 
We can define the loss function of our method as following. 
\begin{equation}
    \begin{split}
        &   \mathcal{L} =    \begin{cases}
    \mathcal{L}_{\text{KD}}^{\text{Annealing}}(i),&\textbf{Stage I: }   1 \le \mathcal{T}_i \leq \tau_{\text{max}}\\
    \mathcal{L}_{CE},              & \textbf{Stage II: }  \mathcal{T}_n = 1
\end{cases} \\
    \end{split}
    \label{eq:annealing1}
\end{equation}
In the above equation, $i$ indicates the epoch index in the training process with the max epoch number of $n$ for stage I, $\mathcal{T}_i$ represents the temperature value at $i^{\text{th}}$ epoch,  $\mathcal{L}_{CE}$ is unchanged from Equation~\ref{eq:KD}, and at each epoch (i), $\mathcal{L}_{\text{KD}}^\text{Annealing}(i)$ is defined as following: 
\begin{equation}
    \begin{split}
        & \mathcal{L}_{\text{KD}}^{\text{Annealing}}(i) = { || {z_s(x)} - {z_t(x)} \times \Phi(\mathcal{T}_i) ||}_2^2 \\
        & \Phi(\mathcal{T}) = 1-\frac{\mathcal{T}-1}{\tau_{\text{max}}}, 1 \le \mathcal{T} \le \tau_{\text{max}}, \mathcal{T}\in \mathbb{N} 
    \end{split}
    \label{eq:annealing2}
\end{equation}
In Equation~\ref{eq:annealing1}, $\mathcal{L}_{\text{KD}}^{\text{Annealing}}$ is defined as an MSE loss between the logits of the student ($z_s(x)$) and an annealed version of the teacher logits ($z_t(x)$), obtained by multiplying the logits by the \textit{annealing function} $\Phi(\mathcal{T})$. The annealing function $\Phi(\mathcal{T})$ can be replaced with any monotonically decreasing function $\Phi:[1, \tau_{\text{max}}] \in \mathbb{N} \rightarrow [0,1]\in \mathbb{R}$. In stage I of our training, initially we set $\mathcal{T}_1 = \tau_{\text{max}}$ (which leads to the most softened version of the teacher outputs because $\Phi(\mathcal{T}_1) = \frac{1}{\tau_{\text{max}}}$) and decrease the temperature during training as the epoch number grows (that is $\mathcal{T}\rightarrow1 \text{ while } i\rightarrow n$). Training in stage I continues until $i=n, \mathcal{T}=1$, for which $\Phi(\mathcal{T}_n) = 1$ and we get the sharpest version of $z_t$ without any softening. The intuition behind using the MSE loss in stage I is that matching the logits of the teacher and student models is a regression task and MSE is one of the best loss functions for this task. We also did an ablation study to compare the performance of MSE and KL-divergence loss function in stage I, and the results of this study support our intuition. For more details, please refer to table \ref{tableAblationLoss:nonlin} of the appendices.

\begin{figure*}[tb]
    \centering
    \includegraphics[width=1.05\textwidth]{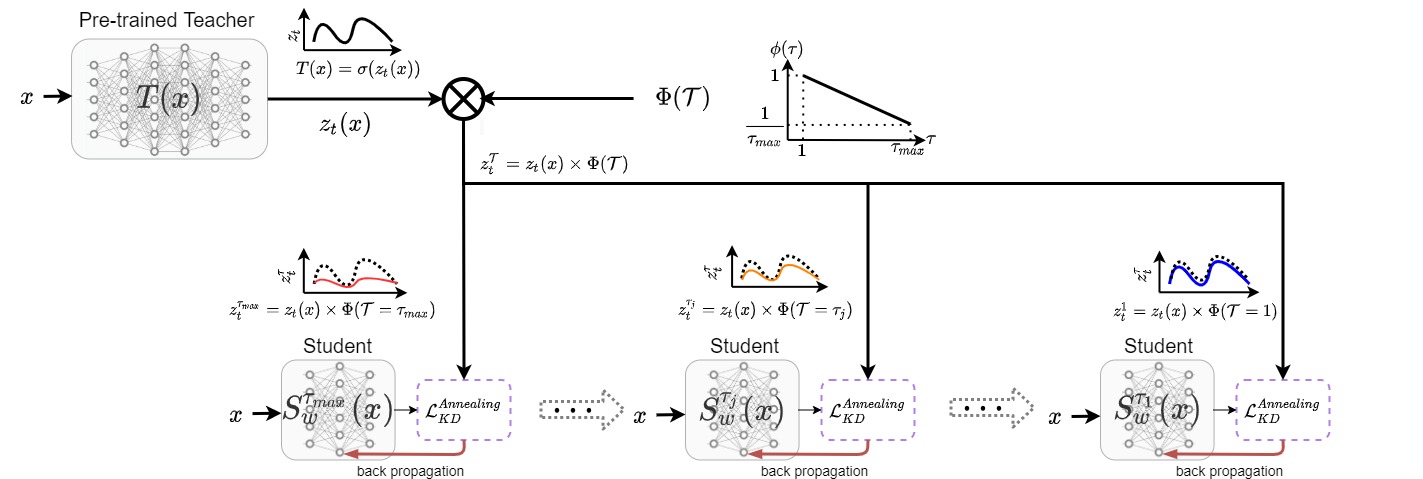}
    \caption{Illustrating the Stage I of the Annealing-KD technique. Given a pre-trained teacher network, we can derive the annealed output of the teacher at different temperature using the annealing function $\Phi(\mathcal{T})$}. We start training of the student from $\mathcal{T}=\tau_{max}$ and go to $\mathcal{T}=1$.
    \label{fig:fig1}
\end{figure*}

Therefore, our Annealing-KD bridges the gap between the student and teacher models by introducing the dynamic temperature term (that is the annealing function $\Phi(\mathcal{T})$) in the stage I of training. This way our Annealing-KD method is able to smoothly transfer the teacher’s knowledge to the student model via a gradual transition over soft-labels generated by the teacher at different temperatures. 

To summarize, our Annealing-KD technique is different from KD in following aspects: 
\begin{itemize}
    \item Annealing-KD does not need any $\lambda$ hyper-parameter to weigh the contribution of the  soft and hard lable losses, because it does the training of each loss in a different stage. 
    \item Our Annealing-KD loss $\mathcal{L}_{\text{KD}}^{\text{Annealing}}$ uses ${||.||}_2^2$ loss instead of the KL divergence. 
    \item Moreover, our technique uses a dynamic temperature by defining the annealing function $\Phi(\mathcal{T})$ in the Annealing-KD loss instead of using a fixed temperature in KD.
    \item Our empirical experiments showed that it is best to take the network logits instead of the softmax outputs in $\mathcal{L}_{\text{KD}}^{\text{Annealing}}$. Furthermore, in contrast to KD, we do not add the temperature term to student output.  
\end{itemize}

Algorithm~\ref{alg} explains the proposed method in more detail.

In this section, we proposed an approach to alleviate the gap between the teacher and student models as well as reducing the sharpness of the KD loss function. 
In our model, instead of pushing the student network to learn a complex teacher function from scratch, we start training the student from a softened version of the teacher and we gradually move toward the original teacher outputs through our annealing process.

\begin{algorithm}[ht]
    \caption{}\label{alg}
    \begin{algorithmic}[1]
        \Function{Annealing-KD}{$S$,$T$,$X$, $k$, $\mathcal{T}_{max}$, $n$}    
            \State 
            
            \Comment{stage I}
            \For{$\mathcal{T} = \tau_{max}$ \texttt{to $1$}}
                
                \State $\Phi \gets 1-\frac{\mathcal{T}-1}{\tau_{\text{max}}}$

                \For{$i = 1$ \texttt{to $k$}}
                    
                    \State \Call{Train-Annealing}{$S$,$T$, $X$,$\Phi$} 
                    \State \Call{Save-Best-Checkpoint}{$S$} 
                \EndFor
            \EndFor
            
            \Comment{stage II}
            \State $S \gets$ \Call{Load-Best-Checkpoint}{}
            
            \For{$i = 1$ \texttt{to $n$}}
                    \State \Call{Train-Fine-Tune}{$S$, $X$} 
                    \State \Call{Save-Best-Checkpoint}{$S$} 
            \EndFor
            
            \State $S \gets$ \Call{Load-Best-Checkpoint}{}

            \State \Return $S$
        \EndFunction
    \end{algorithmic}
\end{algorithm}

\subsection{Example}
\begin{figure}[bt]
 \centering
  \subfloat[]{\label{figur1:1}\includegraphics[width=0.23\textwidth]{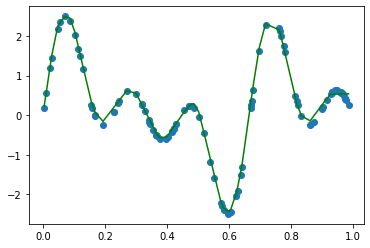}}
  \subfloat[]{\label{figur1:2}\includegraphics[width=0.23\textwidth]{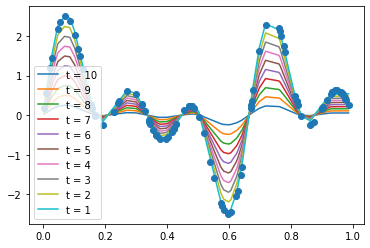}}\\
  \subfloat[]{\label{figur1:3}\includegraphics[width=0.23\textwidth]{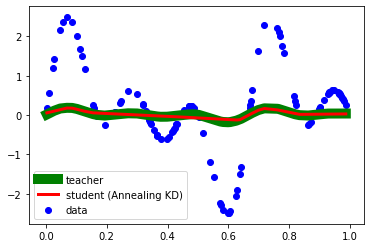}}
  \subfloat[]{\label{figur1:4}\includegraphics[width=0.23\textwidth]{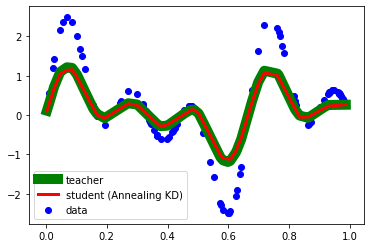}}\\
  \subfloat[]{\label{figur1:5}\includegraphics[width=0.23\textwidth]{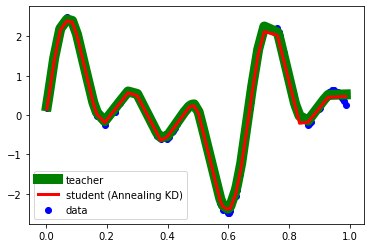}}
  \subfloat[]{\label{figur1:6}\includegraphics[width=0.23\textwidth]{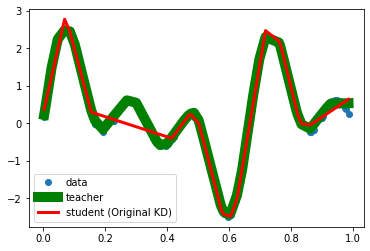}}

  \caption{(a) Data samples and trained teacher. (b) Annealed teacher in different temperatures. (c) Student after matching to annealed teacher in $\mathcal{T}=10$. (d) Student after matching to annealed teacher in $\mathcal{T}=5$. (e) Student after matching to $\mathcal{T}=1$. (f) Student trained without KD.}
  \label{fig2}
\end{figure}

For better illustration of our proposed method, we designed a simple example to visualize different parts of our Annealing-KD algorithm. In this regard, we defined a simple regression task using a simple 2D function. This function is a liner combination of three sinusoidal functions with different frequencies $f(x)=sin(3\pi x)+sin(6\pi x)+sin(9\pi x)$. We randomly sample some points from this function to form our dataset (Figure~\ref{fig2}-(a)). Next, we fit a simple fully connected neural network with only one hidden layer and the sigmoid activation function to the underlying function of the defined dataset. The teacher model is composed of 100 hidden neurons and trained with the given dataset. After training, the teacher is able to get very close to training data (see the green curve in Figure~\ref{fig2}-(a)). We plot the annealed output of the teacher function in 10 different temperatures in Figure.~\ref{fig2}-(b). Then, a student model with 10 hidden neurons is trained once with regular KD (Figure~\ref{fig2}-(f)) and once with our Annealing-KD (Figures.~\ref{fig2}-(c, d, e) depicts the student output at temperatures 10, 5, and 1 during the Annealing-KD training). As it is shown in these figures, Annealing-KD guides the student network gradually until it gets to a good approximation of the underlying function and it can match the teacher output better than regular KD.

\subsection{Rationale Behind Annealing-KD }
Inspired by~\citep{TAKD}, we can leverage the VC-dimension theory and visulaizing loss landscape to justify why Annealing-KD works better than original KD.  
\subsubsection{Theoretical Justification}

In VC-dimension theory \citep{vapnik_1998}, the error of classification can be decomposed as:
\begin{equation}
R(f_s) - R(f) \leq O(\frac{|\mathcal{F}_s|_c}{N^{\alpha_s}}) + \varepsilon_s
\label{vap_1}
\end{equation}

where $R(.)$ is the expected error, $f_s \in \mathcal{F}_s$ is the learner belongs to the function class $\mathcal{F}_s$. $f$ is the underlying function. $|.|_c$ is some function class capacity measure. $O(.)$ is the estimation error of training the learner and $\varepsilon_s$ is the approximation error of the best estimator function belonging to the $\mathcal{F}_s$ class \citep{mirzadeh2019improved}. Moreover, $N$ is the number of training samples, and $\frac{1}{2}\leq \alpha \leq 1$ is a parameter related to the difficulty of the problem.  $\alpha$ is close to $\frac{1}{2}$ for more difficult problems (slow learners) and $\alpha$ is close to 1 for easier problems or fast learners~\citep{lopez2015unifying}. 

In knowledge distillation, we have three main factors: the student (our learner), the teacher, and the underlying function. 
Based on \citep{lopez2015unifying, mirzadeh2019improved}, we can rewrite Equation~\ref{vap_1} for knowledge distillation as following: 
\begin{equation}
R(f_s) - R(f_t) \leq O(\frac{|\mathcal{F}_s|_c}{n^{\alpha_{st}}}) + \varepsilon_{st}
\label{vap_2}
\end{equation}
where the student function $f_s$ is following $f_t$.
To define similar inequality for our Annealing-KD technique, we need to consider the effect of the temperature parameter on the three main functions in KD first. For this purpose, we can define $f_s^\mathcal{T}$, $f_t^\mathcal{T}$, and $f^\mathcal{T}$ as the annealed versions of student, teacher, and underlying functions. Furthermore, let $R_\mathcal{T}(.)$ to be the expected error function w.r.t the annealed underlying function at temperature $\mathcal{T}$. Hence, for Annealing-KD we have 
\begin{equation}
R_\mathcal{T}(f_s^\mathcal{T}) - R_\mathcal{T}(f_t^\mathcal{T}) \leq O(\frac{|\mathcal{F}_s|_c}{n^{\alpha_{st}^\mathcal{T}}}) + \varepsilon_{st}^\mathcal{T}. 
\label{vap_3}
\end{equation}
Note that in $\mathcal{T}=1$, $f_t^1 = f_t$, $f_s^1=f_s$, $f^1=f$, and $R_1(.) = R(.)$. Therefore, we can rewrite Equation~\ref{vap_3} at  $\mathcal{T}=1$ as:
\begin{equation}
R_1(f_s^1) - R_1(f_t^1) \leq O(\frac{|\mathcal{F}_s|_c}{n^{\alpha_{st}^1}}) + \varepsilon_{st}^1.
\label{vap_4}
\end{equation}
That being said, to justify that our Annealing-KD is working better than original KD, we can compare Equations~\ref{vap_4} and ~\ref{vap_2} to show the following inequality holds.   
\begin{equation}
O(\frac{|\mathcal{F}_s|_c}{n^{\alpha_{st}^1}}) + \varepsilon_{st}^1 \leq O(\frac{|\mathcal{F}_s|_c}{n^{\alpha_{st}}}) + \varepsilon_{st}
\label{vap_5}
\end{equation} 

Since in Annealing-KD, the student network at each temperature is initialized with the trained student network at $f_s^{\mathcal{T}-1}$, the student is much closer to the teacher compared with the original KD method, where the student starts from random a initialization. 
In other words, in annealing KD, the student network can learn the annealed teacher at temperature $\mathcal{T}$ faster than the case it starts from a random initial point. Therefore, we can conclude that $\alpha_{st} \leq \alpha_{st}^\mathcal{T}$. This property also holds for the last step of annealing KD where $\mathcal{T}=1$. It means we have $\alpha_{st} \leq \alpha_{st}^1$. Furthermore, bear in mind that since the approximation error depends on the capacity of the learner and in annealing KD we do not change the structure of the student, then we expect to have $\varepsilon_{st} = \varepsilon_{st}^\mathcal{T}$. Therefore, based on these two evidence ( $\alpha_{st} \leq \alpha_{st}^\mathcal{T}$ and $\varepsilon_{st} = \varepsilon_{st}^\mathcal{T}$), we can conclude that Equation~\ref{vap_5} holds.

\subsubsection{Empirical Justification}

Because of the non-linear nature of neural networks, the loss functions of these models are non-convex. This property might prevent a learner from a good generalization. There are some beliefs in the community of machine learning, this phenomena can be harsher in the sharp loss functions than the flat loss functions \citep{chaudhari2019entropy,hochreiter1997flat}. Although, there are some arguments around this belief \citep{li2018visualizing}, for the case of knowledge distillation it seems flatter loss functions are related to higher accuracy \citep{mirzadeh2019improved,zhang2018deep,2015arXiv150302531H}. 
One of the advantages of annealing the teacher function during training is reducing the sharpness of annealing loss function in the early steps of stage I. In other words, the sharpness of the loss function in annealing KD changes dynamically. In the early steps of annealing when the temperature is high, the loss function is flatter. This helps the student to train the teacher network's behaviour faster and easier. 

\begin{figure}[tb]
\centering
    \includegraphics[width=0.5\textwidth]{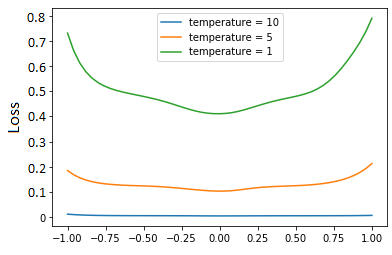}
    \caption{{Visualization of annealing KD loss function in stage I for ResNet 8 student during the training on CIFAR-10 dataset in different temperatures}}
    \label{fig3}
\end{figure}

In order to compare the effect of different temperatures, the loss landscape visualization method in \citep{li2018visualizing} is used to plot the loss behaviour of CIFAR-10 experiment with ResNet 8 student in Figure.~\ref{fig3}. Here as it is shown, by decreasing the temperature during the training, the sharpness of the loss function increases. So the student network can avoid many of the bad local minimums in the early stages of the algorithm when the temperature is high. Then in the final stages of the algorithm, when the loss function is sharper, the network starts from a much better initialization. 





\section{Experiments}
In this section, we describe the experimental evaluation of our proposed Annealing KD method. We evaluate our technique on both image classification and natural language inference tasks. 
In all of our experiments, we compare the annealing KD results with TAKD, standard KD, and training student without KD results. 

\subsection{Datasets}
For image classification, we assess Annealing-KD on CIFAR-10 and CIFAR-100 datasets \citep{krizhevsky2009learning} which are image datasets containing $32 \times 32$ color images with 10 and 100 classes respectively. For the natural language inference task, we employ the General Language Understanding Evaluation (GLUE) benchmark \citep{wang2018glue}, which is a collection of nine different tasks for training, evaluating, and analyzing natural language understanding models. GLUE consists of Multi-Genre Natural Language Inference (MNLI)~\citep{williams2017broad}, Quora Question Pairs (QQP)~\citep{chen2018quora}, Question Natural Language Inference (QNLI)~\citep{rajpurkar2016squad}, Stanford Sentiment Treebank (SST-2)~\citep{socher2013recursive}, Corpus of Linguistic Acceptability (COLA)~\citep{warstadt2019neural}, Semantic Textual Similarity Benchmark (STS-B)~\citep{cer2017semeval}, Microsoft Research Paraphrase Corpus (MRPC)~\citep{dolan2005automatically}, Recognizing Textual Entailment (RTE)~\citep{bentivogli2009fifth}, Winograd NLI (WNLI)~\citep{levesque2012winograd}.

\subsection{Experimental Setup for Image Classification Tasks}
\label{image_classification}
For image classification experiments, we used CIFAR-10 and CIFAR-100 datasets with the same experimental setup in the TAKD method~\citep{TAKD}.  In these experiments, we used ResNet and plain CNN networks as the teacher, student, and also the teacher assistant for the TAKD baseline. For the ResNet experiments, we used ResNet-110 as the teacher and ResNet-8 as the student. For plain CNN experiments, we used CNN network with 10 layers as teacher and 2 layers as the student according to TAKD. Also, for the TAKD baseline, we used ResNet-20 and CNN with 4 layers as the teacher assistant. Tables \ref{tableCV1:nonlin} and \ref{tableCV2:nonlin} compare the annealing KD performance with other baselines over CIFAR-10 and CIFAR-100 datasets respectively. For the ResNet experiments in both tables~\ref{tableCV1:nonlin} and~\ref{tableCV2:nonlin}, the teacher ResNet-110 is trained from scratch and a ResNet-20 TA is trained by the teacher using KD. Then we would like to train a ResNet-8 student using different techniques and compare their performance against our Annealing KD method. In this regard, we evaluate the performance of training the student from scratch, training with the large ResNet-110 teacher using KD, training with TA as the teacher and using our Annealing-KD approach. The results of this experiment with ResNet show that our Annealing-KD outperforms all other baselines and TAKD is the second-best performing student without significant distinction compared to KD. 
More details about the training hyper-parameters are added to the appendix \ref{appendix1}. 


\begin{table}[ht]
\caption{Comparing the test accuracy of annealing KD, TAKD, regular KD, and student without teacher on CIFAR-10 dataset with both ResNet and CNN models } 
\centering 
\resizebox{\columnwidth}{!}{%
\begin{tabular}{c c c c} 
\Xhline{4\arrayrulewidth} 
Model & Type & Training method & Accuracy\\ [0.5ex] 
\hline 
\multirow{6}{*}{ResNet}& Teacher(110) & from scratch & 93.8\\ 
& TA(20) & KD & 92.39\\
\cline{2-4}
& Student(8) & from scratch & 88.44\\
& Student(8) &  KD & 88.45\\
& Student(8) & TAKD & 88.47\\
& Student(8) & \textbf{Annealing KD (ours)} & \textbf{89.44}\\
\hline 
\multirow{6}{*}{CNN} & Teacher(10) & from scratch & 90.1\\ 
 & TA(4) &  KD & 82.39\\
 \cline{2-4}
 & Student(2) & from scratch & 72.75\\
  & Student(2) &  KD & 72.43\\
  & Student(2) & TAKD & 72.62\\
  & Student(2) & \textbf{Annealing KD (ours)} & \textbf{73.17}\\
\hline 
\end{tabular}
}
\label{tableCV1:nonlin} 
\end{table}

\begin{table}[ht]
\caption{Comparing the test accuracy of annealing KD, TAKD, regular KD, and student without teacher on CIFAR-100 dataset with both ResNet and CNN models} 
\centering 
\resizebox{\columnwidth}{!}{%
\begin{tabular}{c c c c} 
\Xhline{4\arrayrulewidth} 
Model & Type &Training method & Accuracy\\ [0.5ex] 
\hline 
\multirow{6}{*}{ResNet} & teacher(110) & from scratch & 71.92\\ 
& TA(20) &  KD & 67.6\\
\cline{2-4}
& student(8) & from scratch & 61.37\\
& student(8) &  KD & 61.41\\
& student(8) & TAKD & 61.82\\
& student(8) & \textbf{Annealing KD (ours)} & \textbf{63.1}\\
\hline 
\multirow{6}{*}{CNN}  & Teacher(10) & from scratch & 64.89\\ 
  & TA(4) &  KD & 60.73\\
  \cline{2-4}
  & student(2) & from scratch & 51.35\\
  & student(2) &  KD & 51.62\\
  & student(2) & TAKD & 51.85\\
  & student(2) & \textbf{Annealing KD (ours)}  & \textbf{53.35}\\
\hline 
\end{tabular}
}
\label{tableCV2:nonlin} 
\end{table}



\begin{table*}[hbt!]
\caption{DistilRoBERTa results for Annealing KD on dev set. F1 scores are reported for MRPC, pearson correlations for STB-B, and accuracy scores for all other tasks.} 
\centering 
\resizebox{\textwidth}{!}{%
\begin{tabular}{c c c c c c c c c c c} 
\Xhline{4\arrayrulewidth} 
KD Method & CoLA & RTE & MRPC & STS-B & SST-2 & QNLI & QQP & MNLI & WNLI & Score\\ [0.5ex] 
\hline 
Teacher & 68.1 & 86.3 & 91.9 & 92.3 & 96.4 & 94.6 & 91.5 & 90.22/89.87 & 56.33 & 85.29\\ 
\hline
From scratch & 59.3 & 67.9 & 88.6 & 88.5 & 92.5 & 90.8 & 90.9 & 84/84 & 52.1 & 79.3\\ 
Vanilla KD & 60.97 & 71.11 & 90.2 & 88.86 & 92.54 & 91.37 & 91.64 & 84.18/84.11 & 56.33 & 80.8\\ 
TAKD & 61.15 & 71.84 & 89.91 & 88.94 & 92.54 & 91.32 & \textbf{91.7} & 83.89/84.18 & 56.33 & 80.85\\
\textbf{Annealing KD} & \textbf{61.67} & \textbf{73.64} & \textbf{90.6} & \textbf{89.01} & \textbf{93.11} & \textbf{91.64} & 91.5 & \textbf{85.34/84.6} & 56.33 & \textbf{81.42} \\ 

\hline 
\end{tabular}
}
\label{table1:nonlin} 
\end{table*}

\tabcolsep=0.11cm
\begin{table*}[ht]
\caption{\small Performance of DistilRoBERTa trained by annealing KD on the GLUE leaderboard compared with Vanilla KD and TAKD. We applied the standard tricks to all 3 methods and fine-tune RTE, MRPC and STS-B from trained MNLI student model. } 
\centering 
\resizebox{\textwidth}{!}{%
\begin{tabular}{c c c c c c c c c c c c} 
\Xhline{4\arrayrulewidth} 
KD Method & CoLA & MRPC & STS-B & SST-2 & MNLI-m & MNLI-mm & QNLI & QQP & RTE & WNLI & Score\\ [0.5ex] 
\hline 
Vanilla KD & \textbf{54.3} & 86/80.8 & 85.7/84.9 & 93.1 & {83.6} & 82.9 & 90.8 & 71.9/89.5 & 74.1 & 65.1 & 78.9 \\
TAKD & 53.2 & 86.7/82.7 & 85.6/84.4 & 93.2 & {83.8} & 83.2 & \textbf{91} & 72/89.4 & \textbf{74.2} & 65.1 & 79 \\
\textbf{Annealing KD} & {54} & \textbf{88.0/83.9} & \textbf{87.0/86.6} & \textbf{93.6} & \textbf{83.8} & \textbf{83.9} & 90.8 & \textbf{72.6/89.7} & 73.7 & 65.1 & \textbf{79.5}\\
\hline 
\end{tabular}
}
\label{table2:nonlin} 
\end{table*}


\begin{table*}[hbt!]
\caption{BERT-Small results for Annealing KD on dev set. F1 scores are reported for MRPC, pearson correlations for STS-B, and accuracy scores for all other tasks.} 
\centering 
\resizebox{\textwidth}{!}{
\begin{tabular}{c c c c c c c c c c c} 
\Xhline{4\arrayrulewidth} 
KD Method & CoLA & RTE & MRPC & STS-B & SST-2 & QNLI & QQP & MNLI & WNLI & Score\\ [0.5ex] 
\hline 
Teacher & 65.8 & 71.48 & 89.38 & 89.2 & 92.77 & 92.82 & 91.45 & 86.3/86.4 & 60.56 & 82.19\\ 
\hline 
Vanilla KD & 33.5 & 57 & 86 & 72.3 & 88.76 & \textbf{83.15} & \textbf{87} & 72.62/73.19 & 54.92 & 70.58\\ 
TAKD & 34.24 & 59.56 & 85.23 & 71.1 & {89.1} & 82.62 & 87 & 72.32/72.45 & 54.92 & 70.76\\
\textbf{Annealing KD} & \textbf{35.98} & \textbf{61} & \textbf{86.2} & \textbf{74.54} & \textbf{89.44} & {83.14} & 86.5 & \textbf{73.85/74.84} & 54.92 & \textbf{71.68} \\

\hline 
\end{tabular}
}
\label{table3:nonlin} 
\end{table*}

\subsection{Experimental setup for GLUE tasks}

For these set of experiments, we use the GLUE benchmark which consists of 9 natural language understanding tasks. In the first experiment (Table \ref{table1:nonlin}), we use RoBERTa-large (24 layers) as teacher, DistilRoBERTa (6 layers) as student, and RoBERTa-base (12 layers) as the teacher assistant for the TAKD baseline. For Annealing KD, we use a maximum temperature of 7, learning rate of 2e-5, and train for 14 epochs in phase 1, and 6 epochs in phase 2. In table ~\ref{table1:nonlin} the Annealing KD and the other baselines performances on dev set of GLUE tasks are compared. Also, we compared the performances of these methods on test set based on the GLUE benchmark's leaderboard results in table ~\ref{table2:nonlin}. In the second experiment (Table \ref{table3:nonlin}), we use BERT-large (24 layers) as teacher, BERT-small (4 layers) as student, and BERT-base (12 layers) as the teacher assistant of TAKD. We use a maximum temperature of 7 for MRPC, SST-2, QNLI, and WNLI, and 14 for all other tasks. The number of epochs in phase 1 is twice the maximum temperature, and 6 in phase 2. We use the learning rate of 2e-5 for all tasks except RTE and MRPC which use 4e-5. Table \ref{table3:nonlin} compares the performance of annealing KD and other baselines on dev set for small-BERT experiments. For more details regarding other hyper-parameters, refer to the appendix. We also perform ablation on the choice of loss function in phase 1, and choice of different max temperature values, both of which can be found in the appendix.

\subsection{GLUE Results}
We present our results in Tables \ref{table1:nonlin}, \ref{table2:nonlin}, and \ref{table3:nonlin}. We see that Annealing KD consistently outperforms the other techniques both on dev set as well as the GLUE leaderboard. Furthermore, in table \ref{table3:nonlin}, when we reduce the size of the student to a 4 layer model (BERT-Small), we notice almost twice as big of a gap in the average score over Vanilla KD when compared with DistilRoBERTa (Table \ref{table1:nonlin}). We can also observe TAKD improving slightly over Vanilla KD, with the improvement being more significant in the case of the smaller student (BERT-Small).

\section{Discussion }
In image classification experiments, the improvement gap between the annealing KD results and the other baselines in CIFAR-100 experiments is larger than CIFAR-10 ones. We can observe similar conditions for the NLP experiments between BERT-small and DistilRoBERTa students (the performance gap of BERT-small is larger). In both of these cases, the problem for the student was more difficult. CIFAR-100 dataset is more complex than CIFAR-10 dataset. So the teacher has learned a more complex function that should be transferred to the student. In NLP experiments, on the other hand, the tasks are the same but BERT-small student has a smaller capacity in compare with DistilRoBERTa. Therefore the problem is more difficult for BERT-small. From this observation, we can conclude, whenever the gap between the teacher and student is larger, the annealing KD performs better than the other baselines and leverage the acquired knowledge by the teacher to train the student.

\section{Conclusion and Future Work}
In this work, we discussed that the difference between the capacity of the teacher and student models in knowledge distillation may hamper its performance.  
On the other hand, in most cases, larger neural networks can be trained better and get more accurate results. If we consider better teachers can train better students, then larger teachers with better accuracy would be more favourable for knowledge distillation training. 
In this paper, we proposed an improved knowledge distillation method called annealing KD to alleviate this problem and leverage the knowledge acquired by more complex teachers to guide the small student models better during their training. This happened by feeding the rich information provided by the teacher’s soft-targets incrementally and more efficiently. Our Annealing-KD technique was based on a gradual transition over annealed soft-targets generated by the teacher at different temperatures in an iterative process; and therefore, the student was trained to follow the annealed teacher output in a step-by-step manner.


\bibliography{references}
\bibliographystyle{apalike}

\newpage

\section*{Appendices}

\appendix
\label{sec:appendix_1}

\section{Experimental parameters of the image classification tasks}
\label{appendix1}
{
In this section, we include more detail of our experimental settings of section \ref{image_classification} in the paper. For the baseline experiments, we used the same experimental setup as \citep{mirzadeh2019improved}. We performed two series of experiments based on ResNet and plain CNN neural networks on CIFAR-10 and CIFAR-100 datasets.  Table \ref{tableCIFAR10:nonlin} illustrates the hyper-parameters used in these experiments. (BS = batch size, EP1= number of epochs in phase 1 (for the baselines, this is the number of training epochs), EP2 = number of epochs in phase 2, LR = learning rate, MO = momentum, WD = weight decay, $\tau_{max}$ = maximum temperature)
}

\begin{table}[h]
\caption{Hyper-parameters of CIFAR-10 and CIFAR-100 experiments} 
\centering 
\resizebox{\columnwidth}{!}{
\begin{tabular}{c c c c c c c c c c} 
\Xhline{\arrayrulewidth} 
Model & Type & Training method & BS & EP1 & EP2 & LR & MO & WD & $\tau_{\text{max}}$ \\ [0.5ex] 
\hline 
\multirow{6}{*}{ResNet}& Teacher(110) & from scratch & 128 & 160 & N/A & 0.1 & 0.9 & 0.0001 & N/A \\ 
& TA(20) & KD & 128 & 160 & N/A & 0.1 & 0.9 & 0.0001 & N/A \\
\cline{2-10}
& Student(8) & from scratch & 128 & 160 & N/A & 0.1 & 0.9 & 0.0001 & N/A \\
& Student(8) &  KD & 128 & 160 & N/A & 0.1 & 0.9 & 0.0001 & 1 \\
& Student(8) & TAKD & 128 & 160 & N/A & 0.1 & 0.9 & 0.0001 & 1 \\
& Student(8) & \textbf{Annealing KD (ours)} & 128 & 160 & 160 & 0.1 & 0.9 & 0.0001 & 10 \\
\hline 
\multirow{6}{*}{CNN} & Teacher(10) & from scratch & 128 & 160 & N/A & 0.1 & 0.9 & 0.0001 & N/A \\ 
 & TA(4) &  KD & 128 & 160 & N/A & 0.1 & 0.9 & 0.0001 & N/A\\
 \cline{2-10}
 & Student(2) & from scratch & 128 & 160 & N/A & 0.1 & 0.9 & 0.0001 & N/A \\
  & Student(2) &  KD & 128 & 160 & N/A & 0.1 & 0.9 & 0.0001 & 1\\
  & Student(2) & TAKD & 128 & 160 & N/A & 0.1 & 0.9 & 0.0001 & 1\\
  & Student(2) & \textbf{Annealing KD (ours)} & 128 & 160 & 160 & 0.1 & 0.9 & 0.0001 & 10\\
\hline 
\end{tabular}
}
\label{tableCIFAR10:nonlin} 
\end{table}

\section{BERT Experiments}
\label{appendix2}
{
In these experiments, RoBERTa-large (24 layers) and DistilRoBERTa (6 layers) are used as the teacher and student models respectively. Also, RoBERTa-base (12-layer) is used as the teacher assistant for the TAKD baseline. For Annealing KD, we use the maximum temperature of 7 and the learning rate of 2e-5 for all the tasks. We trained the student model for 14 epochs in phase 1, and 6 epochs in phase 2. Table \ref{tableDistilRoBERTa:nonlin} illustrates the details of the hyper-parameters of the experiments. 
Also, Table \ref{tableBERTSmall:nonlin} illustrates the hyper-parameter values of BERT-small  experiments in detail. 
Also, we did two ablation studies. In the first one, we tried to fine-tune the maximum temperature in annealing KD and check the performance improvement compared with using the general value of 7. As it is illustrated in Table \ref{tableDistilRoBERTaTemperatureTuning:nonlin}, we can get more improvement with selecting the maximum temperature parameter more carefully. The second ablation is about comparing the effect of mean square error and KL-divergence loss functions on the final results of the experiments when they are used as the loss function of the first phase. Table \ref{tableAblationLoss:nonlin} shows the results of this ablation.
}


\begin{table}[ht]
\caption{Common Hyper-parameters for DistilRoBERTa and BERT-Small models on GLUE tasks} 
\centering 
\resizebox{\columnwidth}{!}{
\begin{tabular}{c c c c c c c c c c} 
\Xhline{\arrayrulewidth} 
Hyper-parameter & CoLA & RTE & MRPC & STS-B & SST-2 & QNLI & QQP & MNLI & WNLI\\ [0.5ex] 
\hline 
Batch Size & 32 & 32 & 32 & 32 & 32 & 32 & 32 & 32 & 32\\ 
Max Seq. Length & 128 & 128 & 128 & 128 & 128 & 128 & 128 & 128 & 128\\
Vanilla KD Alpha & 0.5 & 0.5 & 0.5 & 0.5 & 0.5 & 0.5 & 0.5 & 0.5 & 0.5\\
Gradient Clipping & 1 & 1 & 1 & 1 & 1 & 1 & 1 & 1 & 1\\
\hline 
\end{tabular}
}
\label{tableCommon:nonlin} 
\end{table}

\begin{table}[ht]
\caption{Model specific Hyper-parameters for DistilRoBERTa on GLUE tasks} 
\centering 
\resizebox{\columnwidth}{!}{

\begin{tabular}{c c c c c c c c c c} 
\Xhline{\arrayrulewidth} 
Hyper-parameter & CoLA & RTE & MRPC & STS-B & SST-2 & QNLI & QQP & MNLI & WNLI\\ [0.5ex] 
\hline 
Learning Rate & 2e-5 & 2e-5 & 2e-5 & 2e-5 & 2e-5 & 2e-5 & 2e-5 & 2e-5 & 2e-5\\ 
Phase 1 epochs & 14 & 14 & 14 & 14 & 14 & 14 & 14 & 14 & 14\\
Phase 2 epochs & 4 & 4 & 4 & 4 & 4 & 4 & 4 & 4 & 4\\
$\tau_{max}$ & 7 & 7 & 7 & 7 & 7 & 7 & 7 & 7 & 7\\
\hline 
\end{tabular}
}
\label{tableDistilRoBERTa:nonlin} 
\end{table}

\onecolumn
\begin{table}[ht!]
    \caption{Ablation on DistilRoberta Annealing KD with temperature tuning} 
    \centering 
    \resizebox{0.9\textwidth}{!}{
    \begin{tabular}{c c c c c c c c c c c} 
    \Xhline{\arrayrulewidth} 
    KD Method & CoLA & RTE & MRPC & STS-B & SST-2 & QNLI & QQP & MNLI & WNLI & Avg\\ [0.5ex] 
    \hline 
    {Annealing KD} & {61.67} & {73.64} & {90.6} & {89.01} & {93.11} & {91.64} & 91.5 & {85.34/84.6} & 56.33 & {81.42} \\ 
    \textbf{+ temp tuning} & {61.67} & {73.64} & \textbf{91.99} & \textbf{89.26} & \textbf{93.34} & \textbf{92} & \textbf{91.72} & \textbf{85.14/85.22} & 56.33 & \textbf{81.67} \\ 
    (max temperature) & 7 & 7 & 8 & 14 & 14 & 11 & 14 & 14 & 7 & - \\ 
    
    
    \hline 
    \end{tabular}
    }
    \label{tableDistilRoBERTaTemperatureTuning:nonlin} 

    \vspace{30pt}

    \caption{Ablation on DistilRoberta Annealing KD with different loss functions} 
    \centering 
    \resizebox{0.9\textwidth}{!}{
    \begin{tabular}{c c c c c c c c c c c} 
    \Xhline{\arrayrulewidth} 
    KD Method and Loss & CoLA & RTE & MRPC & STS-B & SST-2 & QNLI & QQP & MNLI & WNLI & Avg\\ [0.5ex] 
    \hline 
    {Annealing KD, MSE} & {61.67} & {73.64} & {90.6} & {89.01} & {93.11} & {91.64} & 91.5 & {85.34/84.6} & 56.33 & {81.42} \\ 
    {Annealing KD, KL-div} & {62.56} & {70.75} & {90.84} & {89.01} & {93} & {91.32} & 91.42 & {85/84.75} & 56.33 & {81.13} \\ 
    
    
    \hline 
    \end{tabular}
    }
    \label{tableAblationLoss:nonlin} 
    
    \vspace{30pt}

    \caption{Model specific Hyper-parameters for BERT-Small on GLUE tasks} 
    \centering 
    \resizebox{0.9\textwidth}{!}{
    \begin{tabular}{c c c c c c c c c c} 
    \Xhline{\arrayrulewidth} 
    Hyper-parameter & CoLA & RTE & MRPC & STS-B & SST-2 & QNLI & QQP & MNLI & WNLI\\ [0.5ex] 
    \hline 
    Learning Rate & 2e-5 & 4e-5 & 4e-5 & 2e-5 & 2e-5 & 2e-5 & 2e-5 & 2e-5 & 2e-5\\ 
    Phase 1 epochs & 28 & 28 & 14 & 28 & 14 & 14 & 28 & 28 & 14\\
    Phase 2 epochs & 6 & 6 & 6 & 6 & 6 & 6 & 6 & 6 & 6\\
    $\tau_{max}$ & 14 & 14 & 7 & 14 & 7 & 7 & 14 & 14 & 7\\
    \hline 
    \end{tabular}
    }
    \label{tableBERTSmall:nonlin} 
    
\end{table}

\end{document}